\documentclass[conference]{IEEEtran}
\IEEEoverridecommandlockouts

\usepackage{cite}
\usepackage{amsmath,amssymb,amsfonts}
\usepackage{algorithm}
\usepackage{algorithmic}
\usepackage{graphicx}
\usepackage{textcomp}
\usepackage{xcolor}
\usepackage{booktabs}

\def\BibTeX{{\rm B\kern-.05em{\sc i\kern-.025em b}\kern-.08em
    T\kern-.1667em\lower.7ex\hbox{E}\kern-.125emX}}
\begin{document}

\title{Toward Collaborative Reinforcement Learning Agents that Communicate Through Text-Based Natural Language}

\author{
\IEEEauthorblockN{Kevin M. Eloff}
\IEEEauthorblockA{\textit{Electrical \& Electronic Engineering} \\
\textit{Stellenbosch University}\\
Stellenbosch, South Africa \\
20801769@sun.ac.za}
\and
\IEEEauthorblockN{Herman A. Engelbrecht}
\IEEEauthorblockA{\textit{Electrical \& Electronic Engineering} \\
\textit{Stellenbosch University}\\
Stellenbosch, South Africa \\
hebrecht@sun.ac.za}
}

\IEEEoverridecommandlockouts
\IEEEpubid{\makebox[\columnwidth]{978-0-7381-1236-7/21/\$31.00~\copyright2021
IEEE \hfill} \hspace{\columnsep}\makebox[\columnwidth]{ }}

\maketitle

\begin{abstract}
Communication between agents in collaborative multi-agent settings is in general implicit or a direct data stream. This paper considers text-based natural language as a novel form of communication between multiple agents trained with reinforcement learning. This could be considered first steps toward a truly autonomous communication without the need to define a limited set of instructions, and natural collaboration between humans and robots. Inspired by the game of Blind Leads, we propose an environment where one agent uses natural language instructions to guide another through a maze. We test the ability of reinforcement learning agents to effectively communicate through discrete word-level symbols and show that the agents are able to sufficiently communicate through natural language with a limited vocabulary. Although the communication is not always perfect English, the agents are still able to navigate the maze. We achieve a BLEU score of 0.85, which is an improvement of 0.61 over randomly generated sequences while maintaining a 100\% maze completion rate. This is a 3.5 times the performance of the random baseline using our reference set.

\end{abstract}


\section{Introduction}
The communication between humans and robots is becoming increasingly relevant in this modern age of machine learning. There have been extensive developments in conversational Artificial Intelligence using state-of-the-art supervised neural models \cite{gao2019neural}. These machine learning techniques allow for a more human-like natural language interaction between man and computer, although they are not generally focused on the actual collaboration of humans and robots. The ability to effectively communicate with a robot through natural language may allow for much greater productivity levels in human-robot interaction, particularly in collaboration. For example, a construction worker could ask an assistant robot to ``Help move this steel beam.'' The robot may respond with ``Okay, where are we moving it to?'' This form of natural language communication could streamline workflow as robots become increasingly autonomous.

In this paper we consider agents trained using reinforcement learning (RL). Specifically, we consider a multi-agent reinforcement learning (MARL) setting where agents must learn to use text-based natural language to effectively collaborate. Communication between agents in collaborative multi-agent settings is in general implicit\footnote{Implicit here refers to the way agents are able to predict the actions of the other agents through observation, without any direct information transmitted.} or a direct data stream\footnote{Direct data streams refers to vectors of continuous values being shared between agents.}. The general goal is to replace this direct data stream with a text-based system and then teach the agents using RL to instead communicate through text-based natural language. This could allow humans to understand the communication used by agents (i.e. robots) and allow for direct human interaction. 

The question we want to answer is whether agents trained by RL are capable of language acquisition. We focus on a simplified problem that considers one-way text communication between two agents. We propose an environment inspired by the game of \textit{Blind Leads}\cite{blindfold}, where one agent instructs another blindfolded agent through some maze. This problem tests both the generation and interpretation of natural language instructions by the respective collaborating agents. 

In one agent, we combine a convolutional neural network (CNN) with a recurrent neural network using a gated recurrent unit (GRU) architecture. A CNN encodes the environment, which is then decoded into an instruction by a GRU. The second agent makes use of two GRUs to encode and decode this instruction into an action sequence.

\section{Related Work}

Due to recent technological advances in computing power and the increasing popularity of machine learning, recent papers have been made in the field of RL based language acquisition. The following papers have both shown the emergence of language in MARL.

\subsection{Emergent Communication in Multi-agent Navigation}
Kaji\'{c} et al. \cite{kajic+etal_2020} investigated a similar setup to the one considered in this paper. They also investigate interactive MARL setups where agents are allowed to communicate through signals while in an environment. Kaji\'{c} wanted to investigate the relationship between the environment layout as well as the discrete signals transmitted between agents. They considered emergent communication between RL agents in the context of co-operative games.  

Kaji\'{c} proposed a game in the form of a navigation task. They tasked an agent with navigating to a specific unknown goal in a set environment. This agent relied on signals from a separate observing agent which knows the goal location. The agent is set in a bounded 2-dimensional environment. At each step, the receiver agent is presented with the environment state and an encoded goal signal constructed by the sender agent. The receiver agent then chooses an action, being one of the 4 cardinal directions (North, South, East, or West). The sender agent is given a representation of the environment along with the goal location from which it generates the encoded signal to be interpreted by the receiver. 

They demonstrated the capability of agents to effectively communicate and studied the emergent communication protocols generated by these agents. A non-communicating Q-learning agent that may observe the goal was used as a baseline. They found that the communicating agents had comparable performance to that of a single Q-learning agent. Kaji\'{c} was thus able to prove the capability of RL in the task of language acquisition given a simple communication task.

Although Kaji\'{c} did show a promising avenue of emergent communication within multi-agent RL, they did not attempt to ground the communication protocol to any specific existing human language (such as English). Rather they let the agents create languages of their own.

\subsection{Emergence of Language with Multi-agent Games}

As an alternative to the environment proposed previously, Havrylov and Titov \cite{havrylov+titov_2017} propose a different approach with a similar goal. They also recognised the potential of learning to communicate through interaction, rather than learning through supervised methods. Their goal was to assess the emergent communication between agents while constraining communication to have properties similar to that of natural language.

Havrylov and Titov propose a referential game environment where agents are required to communicate through discrete messages in the form of a language. Similar to the setup of Kaji\'{c}, the environment has both a sender and a receiver agent. The sender agent is presented with an image and tasked with generating a message. This message is then presented to the receiving agent along with three images, one of which is the same as the image presented to the receiver agent. The goal of the agents is for the receiver agent to correctly identify the image presented to the sender agent. The agents are built using CNNs and Long Short-Term Memories (LSTMs). 

They compared an RL approach to a differentiable approach using a Gumbel-softmax estimator. They found that both solutions result in effective communication protocols, with the differentiable relaxation being faster to converge. They found the learned communication protocol to have characteristics of natural language. They also investigated methods of grounding presenting the agents with prior information about a preexisting natural language. 

\cite{havrylov+titov_2017} is more aligned to our goals than \cite{kajic+etal_2020}, as \cite{havrylov+titov_2017} had a greater focus on grounding the emergent communication protocol to a natural language (i.e.~English).

\section{Proposed Solution}

We now describe the proposed environment in which we would like to test the effectiveness of our learned communication protocol, as well as the proposed solutions and baseline. This environment takes the form of a referential game inspired by a classic team-building exercise commonly known as \textit{Blind Leads}. This game generally involves one blindfolded person and one guide, where the goal is for the guide to direct the blindfolded solely via communication in order to navigate some environment or to complete some task. This leads us to our adaptation of the exercise.

\subsection{Environment}
The environment will take the form of a 2-dimensional maze, where one agent is placed randomly in the environment. This agent, known as the \textit{receiver}, is analogous to the blindfolded person mentioned previously. The receiver is unable to observe the maze, they are only allowed to execute actions within the environment. The other agent, known as the \textit{sender}, is allowed to observe the maze. The sender agent must communicate with the receiver agent for it to correctly navigate the maze and reach the goal. 

\begin{figure}[htbp]
    \centering
    \includegraphics[width = 1\linewidth]{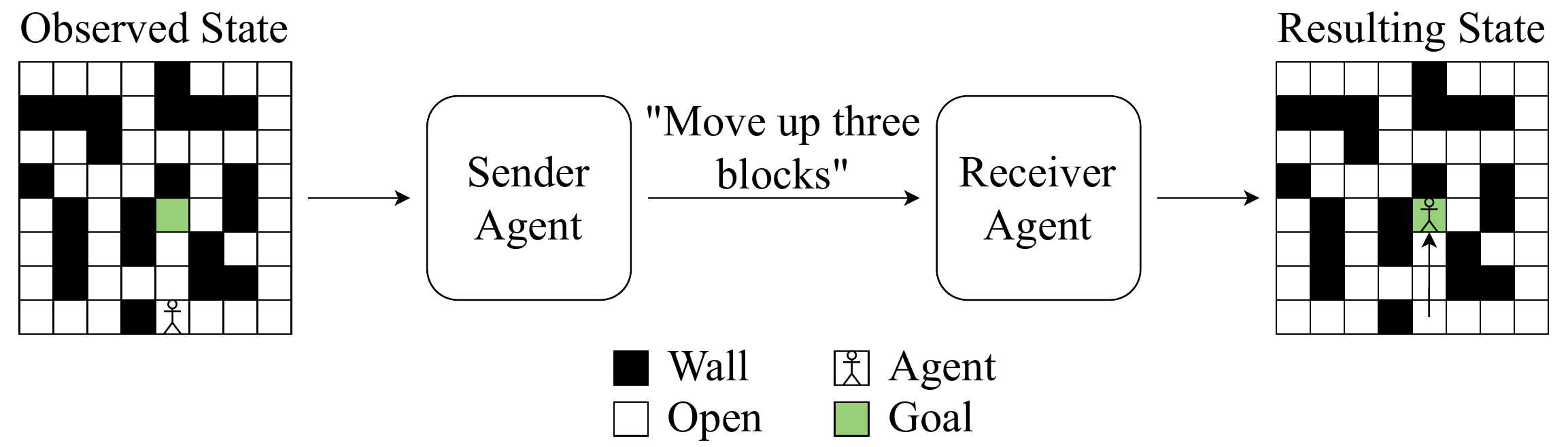}
    \caption{Example interaction between each agent and the environment.}
    \label{fig:2_1a_example}
\end{figure}

Fig.~\ref{fig:2_1a_example} shows an example interaction between each agent and the environment. The sender agent first observes the environment. This agent notes that for the receiver agent to reach the goal, it needs to move north. The sender agent, therefore, generates the sentence ``Move up three blocks''. The receiver agent then decodes this sentence and generates an action sequence, moving it north three blocks towards the goal. The cycle is then repeated, where the sender agent observes the resulting state and generates a new instruction.

The communication between the agents will consist of a sequence of discrete word-level symbols with maximum length $M$. Each symbol will link to a word in a vocabulary of size $V$. The receiver agent is allowed to execute a maximum of $N$ actions at each environment step.

\subsection{Sender Agent}

The goal of the sender agent is to generate a sequence of discrete symbols, or message, given some observed environment state. This message should contain information instructing the receiver agent on how to move to reach the final goal position. We can therefore split this agent up as an \textit{encoder-decoder}\cite{seq}. First, the sender agent encodes the environment, capturing any information regarding what sort of actions the receiver agent would need to take to reach the goal. This encoding is then decoded into a message sequence describing in words what actions the receiver should take.

\subsubsection{Environment State Encoder}
\begin{figure}[htbp]
    \centering
    \includegraphics[width = 1\linewidth]{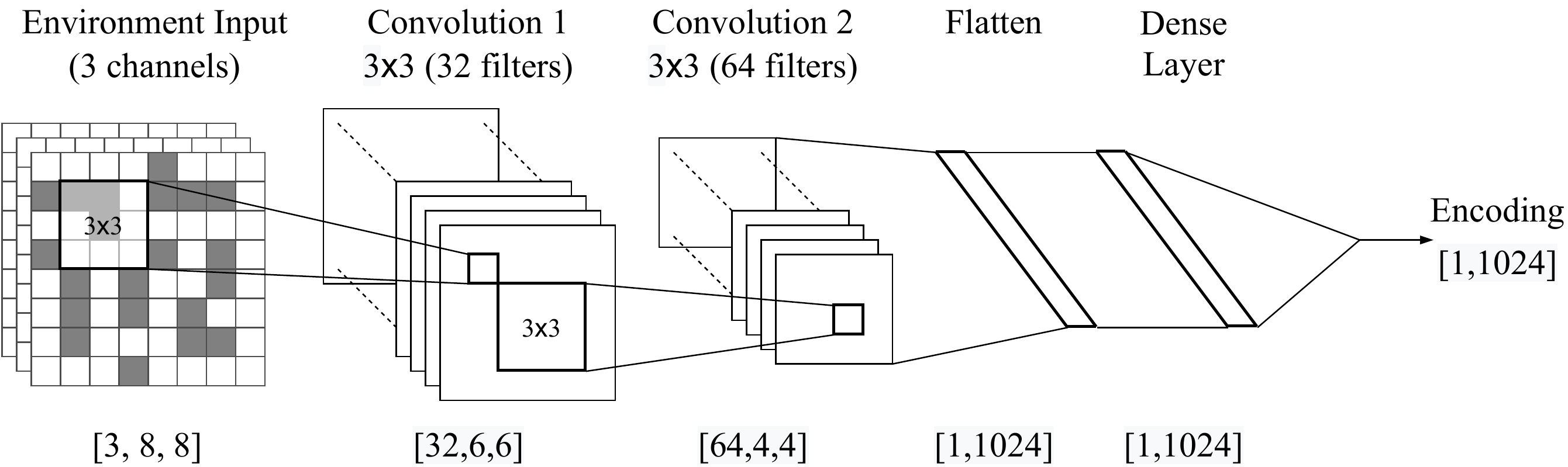}
    \caption{State encoding convolutional layers.}
    \label{fig:2_2a_sender_conv}
\end{figure}

To encode the environment state, we make use of two convolutional layers and one dense layer. We also represent the environment state as $3$ separate channels. We have a separate channel representing the maze, receiver agent position, and goal position. Each convolutional layer makes use of $3\times3$ filters with a stride of 1. The first and second layer have 32 and 64 filters respectively. The output of the second convolutional layer is flattened into a $1\times1024$ vector which is fed through a single dense layer. The output of this dense layer is the encoded output. Fig.~\ref{fig:2_2a_sender_conv} shows how each layer of the encoding section is connected. 

\subsubsection{Message-Generating Decoder}
From the encoder mentioned previously, we now need to implement the decoder. The decoder should be able to generate an arbitrary-length message sequence given some vocabulary. To do this, we require some sort of recurrent memory and therefore chose to use a GRU. The GRU, first proposed by Cho et al. \cite{cho2014learning}, has been proven to have performance equivalent to LSTMs while also being much easier to train \cite{chung2014empirical}. Both LSTMs and GRUs are capable of retaining long term information unlike standard RNNs, thus making them ideal for use in sequence generation and sequence-to-sequence models.

\begin{figure}[htbp]
    \centering
    \includegraphics[width = 1\linewidth]{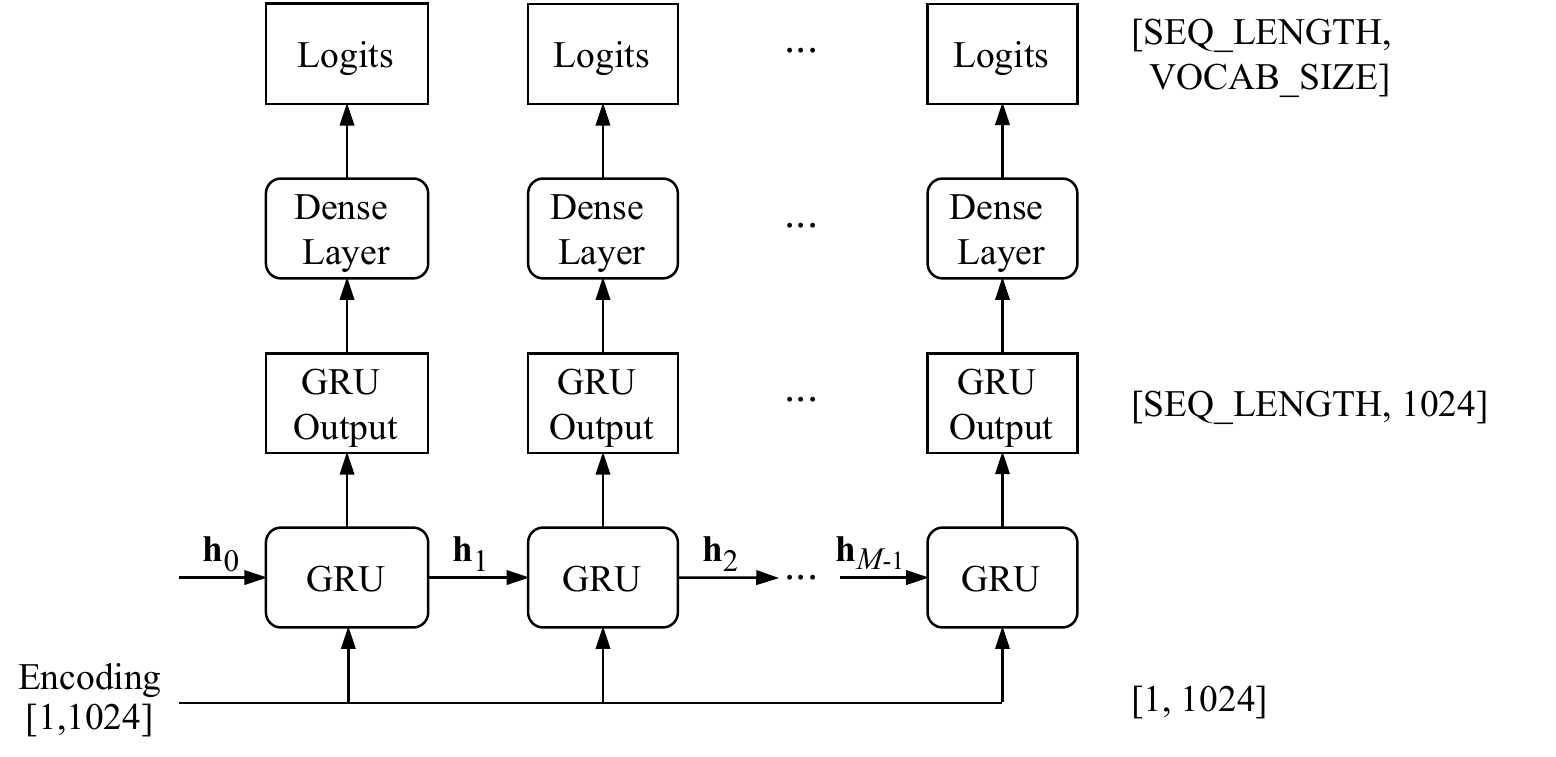}
    \caption{Sequence generating GRU Decoder.}
    \label{fig:2_2b_sender_gru}
\end{figure}

By using a GRU in our decoder setup, we input the encoding of the previous step directly as the input to each unit as shown in Fig.~\ref{fig:2_2b_sender_gru}. The initial hidden state\footnote{The hidden states referred to throughout this paper are not shared between each GRU, $\textbf{h}$ refers to each respective GRU separately} $\textbf{h}_0$ is initialised to zeros. We apply a dense layer and soft-max to the output of each unit. The purpose of this layer is to convert each $1024$ length GRU output into the final one-hot encodings representing words from our vocabulary. The full sequence of these words forms the message generated by the sender agent.

\subsubsection{Connected Encoder-Decoder}
By combining the encoder and decoder, we create a model capable of interpreting some environment state and generating a sequence of symbols. We note that the combination of the two is also end-to-end differentiable, which is required when training. Due to the convolution layers used at the input, the sender agent should be able to better generalise and adapt to changes in the map environment. The GRU at the output also allows for reasonably complex arbitrary length sequences.

\subsection{Receiver Agent}

The receiver agent is slightly simpler than the sender agent. The goal of the receiver agent is to generate an action sequence given a sequence of discrete symbols. This sequence of discrete symbols is the message generated by the sender agent. We therefore view this receiver agent as a sequence-to-sequence network. The agent needs to encode a sequence, or message, into an internal state representation capturing any information regarding which actions it should take. This is then decoded into an action sequence or action plan that this agent will execute in the environment. 

\subsubsection{Message Encoder}
\begin{figure}[htbp]
    \centering
    \includegraphics[width = 1\linewidth]{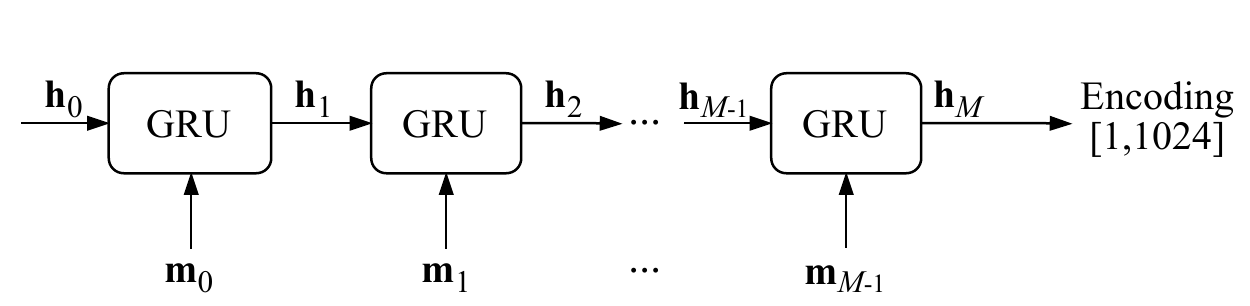}
    \caption{GRU message decoder.}
    \label{fig:2_3a_receiver_gru1}
\end{figure}

The encoder must transform the message sequence into an internal state representation. We, therefore, make use of a GRU due to its ability to retain information in arbitrary length sequences. Assume the sender agent generated a sequence of symbols defined as the message $\textbf{m}$ of length $M$. This message is interpreted by a GRU as shown in Fig.~\ref{fig:2_3a_receiver_gru1}. The input size of each unit is equal to the vocabulary size used in the message, as we make use of one-hot encoding.

We initialise the initial hidden state $\textbf{h}_0$ as a 1024 length zero vector. The final hidden state of the GRU is used as the internal encoding. The outputs are discarded.

\subsubsection{Action Sequence Generating Decoder}
The decoder used in the receiver agent is very similar to that of the sender agent shown in Fig.~\ref{fig:2_2b_sender_gru}. The only difference being the output size, as it now reflects the total number of possible actions. A similar softmax is used to select a one-hot encoding representing the selected action. It is important to note that the GRU used in the decoder is different from the one used here, thus the sender agent consists of two separate GRUs.

Combined, the encoder and decoder form a sequence-to-sequence model whereby we can generate an action sequence given a natural language message. We note this combination of the two GRUs is also end-to-end differentiable.

\subsection{Baseline}
The baseline aims to provide a separate simpler solution to compared our proposed solutions. Our baseline comes in two forms, the first being a single agent that mimics the behaviour of the combined sender and receiver agents without communication. The second baseline is a tiered algorithm approach that makes use of hard-coded methods to split up the overall problem.

\subsubsection{Single Agent}
The single agent is constructed through the combination of the sender and receiver agent and then removing the communication segments. This may also be seen as an encoder-decoder network. The goal of this agent is to encode the environment state and then decode it into an action sequence. We use the environment state encoder of the sender agent and the action sequence generating decoder of the receiver agent. Theoretically, if the communication between the sender and receiver is perfect, the performance of the combination should be equal to that of the single agent.

\subsubsection{Tiered Algorithm}
To create the hardcoded solution, we break the problem space up into three segments: (1) state to action sequence, (2) action sequence to discrete message, and (3) discrete message to action sequence. The first two mirror the functionality of the sender agent, and the third mirrors that of the receiver agent. 

\par\noindent (1) The state to action sequence algorithm generates an action sequence given the environment state. This translates to a list of actions describing the optimal path towards the goal from the current position. To solve the maze, we convert the output of an A* algorithm into an action sequence.

\par\noindent (2) The action sequence to discrete message algorithm is relatively simple and generates a text sentence or message given a list of actions. The algorithm makes use of random selection to generate a description of the desired action sequence.

\par\noindent (3) The discrete message to action sequence algorithm interprets a natural language message and generates an action sequence thereof. We make use of Levenshtein Distance to map the message sequence to a set of known messages and their respective actions. From \cite{Schulz,Navarro}, we define the Levenshtein distance between two strings as the number of editing operations (deletions, insertions, substitutions) required to convert one sentence to another. This allows for slightly more freedom in the message format while still having a reasonably good interpretation.

\section{Experimentation Setup}

To train our agents, we will make use of an adaption of the deep Q-learning algorithm proposed by Mnih et al. \cite{mnih+etal_2013}. From Mnih et al. Q-learning is an RL algorithm with the goal of finding the best action given the current state. The agent is able to make an observation $s_t$ of the environment \(\mathcal{E}\)\ state at each time-step $t$ and selects an action $a_t$ from a set of legal environment actions \(\mathcal{A} = \{1,...,K\}\). The agent executes the action in the environment and is presented with a scalar reward $r_t$. The environment \(\mathcal{E}\)\ is then updated and the new state is once again observed by the agent. Each cycle is a round and the cycle is repeated a finite amount until an end condition is met. Q-learning seeks to maximise the total reward over all rounds. The reward is generally set to encourage the agent to solve some task or reach some goal within the environment. This translates selecting an action that maximises the action-value function $Q(s,a)$ according to the Bellman equation\cite{bellman_1954}.

\subsection{Game Environment Setup}
\begin{figure}[ht]
    \centering
    \includegraphics[width = 0.8\linewidth]{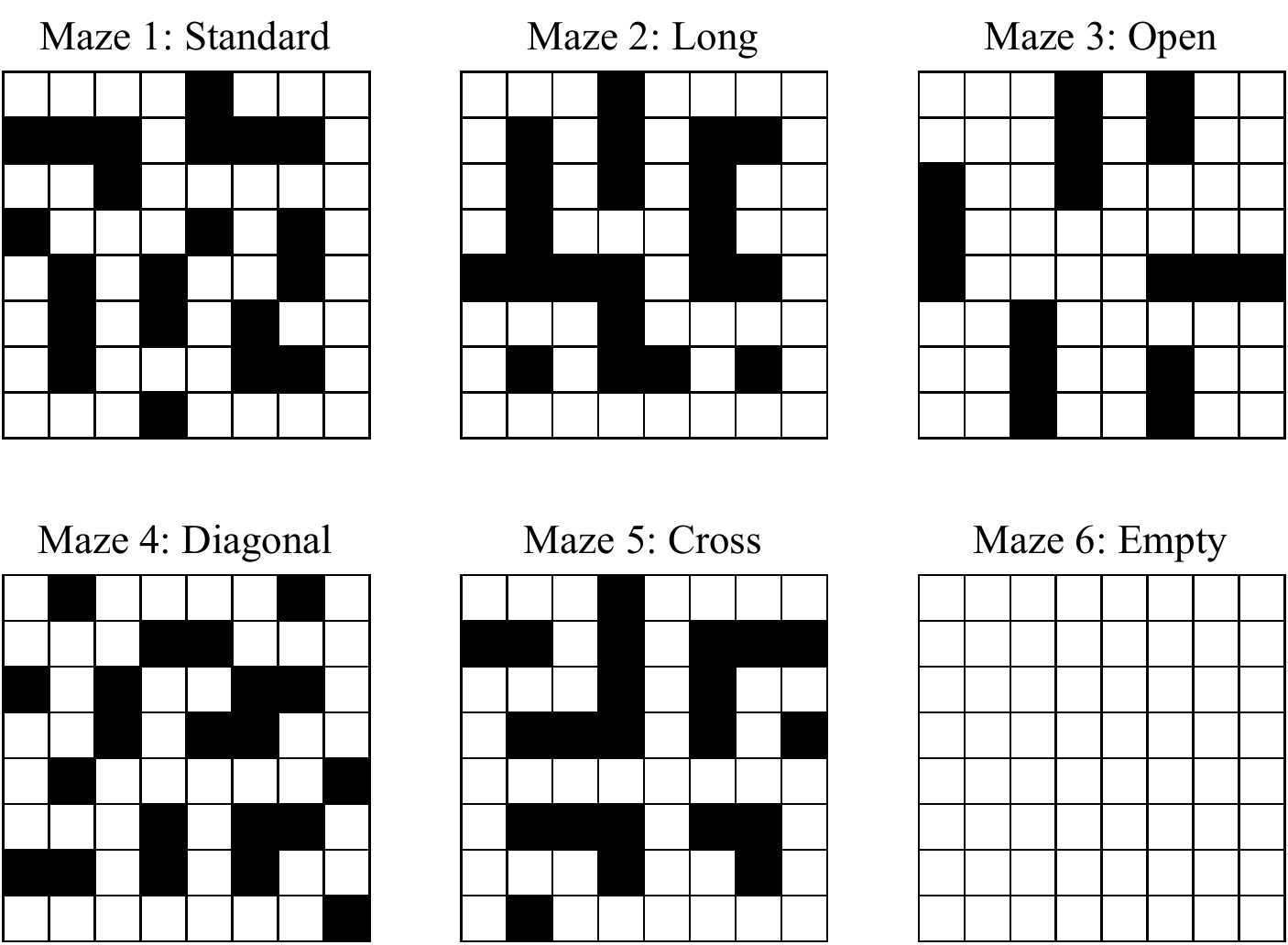}
    \caption{Visualisation of each maze.}
    \label{fig:2_1a_mazes}
\end{figure}

A discrete set of 6 mazes will be used as the environment. Each maze will be represented using a 2-dimensional $8\times8$ float vector, where $0.0$ indicates open space and $1.0$ indicates a wall. A visualisation of the set of mazes is shown in Fig.~\ref{fig:2_1a_mazes}. Within the maze environment, the agent is allowed to take a standard set of Cartesian co-ordinate actions. We define the set of possible actions $\textbf{A}$ where
\vspace{-0.05cm}
\[ \textbf{A} = \{\texttt{NORTH}, \texttt{ SOUTH}, \texttt{ EAST}, \texttt{ WEST}, \texttt{ NONE}\}.\]
\vspace{-0.5cm}

The typical game cycle follows that of a standard Q-learning environment, where the agent observes the state $s_t$ of the environment and generates some action $a_t$. In our case, the environment allows the execution of a sequence of actions at each round\footnote{In our environment, a round refers to a single observation-action cycle while a step refers to an action movement. Therefore multiple steps may be taken per round. An episode is a set of rounds until the maze is solved, or the maximum steps is reached.}. This is to encourage more detailed action descriptions.

The communication protocol within the environment is limited to a 4 length sequence of one-hot encodings. These encodings represent items from a limited vocabulary. The vocabulary itself is limited to 31 English words relating to simple commands such as ``Move up three blocks''.

\subsection{Training Algorithm}
In order to solve the environment, we need to estimate the action-value function. We define a neural network action-value approximator with weights $\theta$ as a Q-network. We may train this Q-network by minimising a set of loss functions $L_i(\theta_i)$,

\begin{equation}\label{equ:q_loss_func}
\begin{split}
L_i(\theta_i) & = \mathbb{E}_{s,a\sim\rho(\cdot)}\big[(y_i - Q^*(s,a;\theta_i))^2\big]
\end{split}
\end{equation}
\vspace{-0.4cm}
\begin{equation}
\begin{split}
y_i & = \mathbb{E}_{s'\sim\mathcal{E}}\big[r + \gamma \max_{a'} Q^*(s',a';\theta_{i-1})\big|s,a\big]
\end{split}
\end{equation}

Here, $y_i$ is target action-value when action $a$ is taken at state $s$ for iteration $i$ and $\gamma$ is the discount for future reward. $\rho(s,a)$ is behavior distribution which gives a probability distribution over actions $a$ and states $s$. We fix the model parameters $\theta_{i-1}$ of the previous iteration when optimising the loss function $L_i(\theta_i)$. It is interesting to note that the targets $y_i$ actually depend on the network weights. This is a main difference between deep reinforcement learning and standard supervised learning, where the targets are generally fixed before training. In order to train the network, we first need to differentiate the loss function with respect to the weights to obtain the gradient:

\begin{equation}\label{equ:derived_ov_gradient}
\begin{split}
\nabla_{\theta_i} L_i(\theta_i) = \mathbb{E}_{s,a\sim\rho(\cdot);s'\sim\mathcal{E}} \big[ & \big(r + \gamma Q^*(s',a'; \theta_{i-1}) \\&- Q^*(s,a;\theta_i)\big) \nabla_{\theta_i}Q^*(s,a;\theta_i)\big]
\end{split}
\end{equation}

We complement the deep Q-learning algorithm through the use of a target network and experience replay from \cite{mnih2015humanlevel}. The final training algorithm is as follows:

\begin{algorithm}[H]
\small
\caption{Deep Q-learning with Experience Replay and target Q-network}
\label{alg1}
\begin{algorithmic}
\STATE Initialise replay memory $\mathcal{D}$
\STATE Initialise target action-value function $Q_{target}$ with random $\theta$
\STATE Initialise policy action-value function $Q_{policy}$ with random $\theta$
\FOR{episode = 1, $N$}
\STATE Reset environment $\mathcal{E}$
\STATE Update $\epsilon$ as a function of $N$
\FOR{$t=1$, $T$}
\STATE With probability $\epsilon$, select random action sequence $\textbf{a}_t$
\STATE otherwise select $\textbf{a}_t = \text{max}_\textbf{a}Q^*_{policy}(\textbf{s}_t, \textbf{a};\theta)$
\STATE $\textbf{s}_{t+1}, r_t \leftarrow$ Execute $\textbf{a}_t$ in environment $\mathcal{E}$ and observe
\STATE Store $(\textbf{s}_t, \textbf{a}_t, r_t, \textbf{s}_{t+1})$ in $\mathcal{D}$
\STATE Sample random batch of transitions $(\textbf{s}_j, \textbf{a}_j, r_j, \textbf{s}_{j+1})$ from $\mathcal{D}$
\STATE $y_j \leftarrow  \begin{cases}r_j & \text{for terminal } \textbf{s}_j\\r_j + \gamma\ \text{max}_\textbf{a}Q_{target}(\textbf{s}_j, \textbf{a};\theta) & \text{for non-terminal } \textbf{s}_j\end{cases}$
\STATE Perform a gradient descent step on $Q_{policy}$ according to equation~\ref{equ:derived_ov_gradient}
\ENDFOR
\STATE Periodically update $Q_{target}$ with $Q_{policy}$ every $M$ episodes
\ENDFOR
\end{algorithmic}
\end{algorithm}

This algorithm is used to train each agent on the 6 mazes. The agents are initially trained in isolation using the tiered algorithms mentioned previously. For the receiver agent, we use the tiered algorithm to generate action instruction given the environment. This instruction is then processed by the receiver agent to generate an action sequence. To train the sender agent, we task it with generating an instruction message given the environment. This message is then interpreted by our tiered algorithm to generate an action sequence. The baseline agent is tasked with directly generating an action sequence given the environment.
The reward function of each agent is the same. The agent is rewarded -0.04 for each step taken, -0.75 for an invalid action, -0.25 for moving to a block previously seen, and +1 for reaching the goal.

\section{Results}

We first trained each agent individually on a per-map basis with a set goal position of $(4,4)$ on each map until convergence. This was done to speed up training. Although training on all maps with random goal positions is also possible. We made use of the tiered algorithm to assist in training the sender and receiver agents prior to combining them. 

Once trained, the solutions were tested on each maze with $\epsilon = 0.01$ for 2000 episodes. Each episode is limited to a maximum of 100 steps, where if reached, the episode counts as a loss. Tab.~\ref{tab:results} shows the maze completion rate of each solution. In this table, Random refers to an agent where a random valid action is taken in each round. TR and SA refer to the tiered and single agent baselines respectively. Combo refers to the combination of the pre-trained sender and receiver agent, and T-Combo refers to the same combination although the receiver agent has been trained on the sender agent's output. 

\begin{table}[htbp]
\caption{Maze completion rate of each solution per map}
\begin{center}
\scriptsize
\begin{tabular}{c c c c c c c c}
Maze\# & Random & TR & SA & Sender & Receiver & Combo & T-Combo\\
\toprule
1 & 0.59 & 1.00 & 1.00 & 0.96 & 1.00 & 0.75 & 1.00 \\
2 & 0.55 & 1.00 & 1.00 & 0.95 & 1.00 & 0.92 & 1.00 \\
3 & 0.72 & 1.00 & 1.00 & 0.96 & 1.00 & 0.84 & 1.00 \\
4 & 0.66 & 1.00 & 1.00 & 0.89 & 1.00 & 0.70 & 1.00 \\
5 & 0.77 & 1.00 & 1.00 & 0.92 & 1.00 & 0.68 & 1.00 \\
6 & 0.69 & 1.00 & 1.00 & 0.94 & 1.00 & 1.00 & 1.00 \\
\end{tabular}
\label{tab:results}
\end{center}
\end{table}

We note that the sender agent on its own could not reach a 100\% win rate, unlike the trained combination. The receiver agent is able correct the flaws in the sender agent's message generation, although this does require further training.

The average number of rounds taken by each solution was also calculated as shown in Tab.~\ref{tab:results_2}. This indicates how efficiently the agent was able to solve the maze, where lower is better. We note that the trained sender-receiver combination is able to reach similar efficiency to that of the baseline. We also note that although the win-rate of the untrained combination is relatively low, the average rounds taken is still similar to that of the standalone sender agent.

\begin{table}[htbp]
\caption{Average rounds taken by each solution per map}
\begin{center}
\scriptsize
\begin{tabular}{c c c c c c c c}
Maze\# & Random & TR & SA & Sender & Receiver & Combo & T-Combo \\
\toprule
1 & 71.2 & 3.58 & 3.83 & 8.59 & 5.02 & 10.59 & 3.45 \\
2 & 66.3 & 3.32 & 3.17 & 8.81 & 4.36 & 11.25 & 3.01 \\
3 & 54.8 & 2.39 & 2.31 & 6.92 & 3.22 & 6.97 & 2.48 \\
4 & 57.2 & 3.59 & 3.53 & 14.91 & 4.7 & 10.86 & 3.65 \\
5 & 49.0 & 2.15 & 2.51 & 13.13 & 3.18 & 11.25 & 2.17 \\
6 & 58.5 & 1.8 & 2.51 & 10.89 & 2.66 & 2.79 & 1.98
\end{tabular}
\label{tab:results_2}
\end{center}
\end{table}

We use the \textit{BLEU} scoring metric\cite{bleu} to measure the quality of the learned communication. The weighting of the BLEU scoring used is 0.75 and 0.25 for unigram and bigram respectively. The reference set of the BLEU metric is generated by recording the unique sentences from 10000 episodes of the tiered algorithm. We score the sentences generated in the previous tests and compare them to those of randomly generated sequences. The results are shown in Tab.~\ref{tab:bleu}. We observe an average improvement of around 0.61 in the BLEU score of our solution to that of randomly generated sequences. The BLEU score of the trained combination is relatively high as expected due to the limited vocabulary and simple instruction set.

\begin{table}[htbp]
\caption{BLEU score of generated sequences to reference set}
\begin{center}
\scriptsize
\begin{tabular}{c c c}
Maze\# & Random & T-Combo \\
\toprule
1 & 0.246 & 0.801 \\
2 & 0.241 & 0.888 \\
3 & 0.250 & 0.837 \\
4 & 0.235 & 0.852 \\
5 & 0.237 & 0.835 \\
6 & 0.244 & 0.904 \\
\midrule
Average: & 0.242 & 0.853 \\
\end{tabular}
\label{tab:bleu}
\vspace{-0.2cm}
\end{center}
\end{table}

\begin{figure}[ht]
    \centering
    \includegraphics[width = 0.9\linewidth]{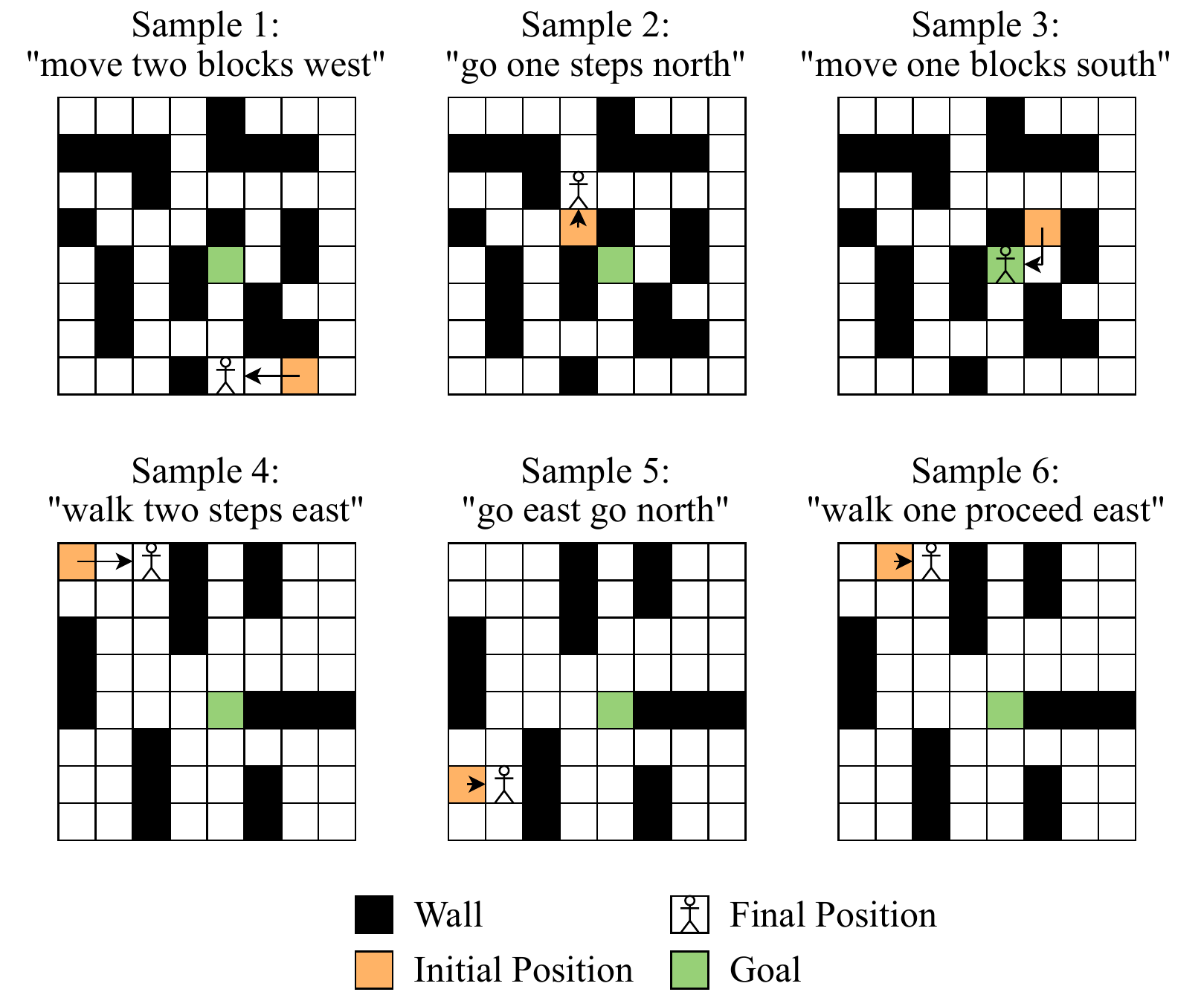}
    \caption{Random samples of the agent communication.}
    \label{fig:4_1a_samples}
\end{figure}

Additionally, we performed qualitative analysis on the communication that was learned. Fig.~\ref{fig:4_1a_samples} shows random samples generated by the trained sender-receiver combination. Each sample shows the initial receiver agent position, the instruction generated by the sender agent, and the final position of the receiver agent after action execution.

We see that the sender agent is able to generate English instructions although the grammar is not always perfect. The receiver agent is generally able to correctly interpret the instruction. We also note that the receiver agent occasionally finds exploits as depicted in Sample 3. In this case, the agent moves down and left to reach the goal rather than following the instruction and only moving down one block.

\section{Conclusion}

In conclusion, we were able to develop models for sender and receiver agents by combining CNNs and GRUs in encoder-decoder formats.
We have shown that these agents are able to sufficiently communicate through natural language with a limited vocabulary. They are able to reach similar performance to that of a baseline which excludes any communication. Although the communication is not always perfect English, the agents are still able to effectively navigate the maze. 

We have shown that agents in such a setting are capable of learning language through RL. Our combined sender-receiver achieved a BLEU score of 0.853, which is an improvement of 0.61 over randomly generated sequences while still maintaining a 100\% maze completion rate.

For future work, we will consider a more complex environment, allowing for more detailed instructions. We will also consider using a pre-trained model to evaluate the grammar and incorporate it into the reward function. As a further extension, we intend on using continuous speech signals rather than discrete word-level symbols. We also intend on investigating two-way communication, which would introduce a more conversational aspect between agents.

\bibliography{conference_101719}

\end{document}